\documentclass{article}

\usepackage{microtype}
\usepackage{graphicx}
\usepackage{subfigure}
\usepackage{booktabs}
\usepackage{hyperref}

\usepackage[accepted]{icml2024}

\usepackage{amsmath}
\usepackage{amssymb}
\usepackage{mathtools}
\usepackage{amsthm}
\usepackage[capitalize,noabbrev]{cleveref}

\icmltitlerunning{DepthWeave-KV: Token-Adaptive Cross-Layer Residual Factorization for Long-Context KV Cache Compression}

\begin{document}

\twocolumn[
\icmltitle{DepthWeave-KV: Token-Adaptive Cross-Layer Residual Factorization for Long-Context KV Cache Compression}

\begin{icmlauthorlist}
\icmlauthor{Anna C\'ordoba}{ivar}
\icmlauthor{Adam Puente Tercero}{ivar}
\icmlauthor{Nerea Angulo Hijo}{ivar}
\icmlauthor{Mar Linares Tercero}{ivar}
\icmlauthor{Julia Barrientos}{ivar}
\icmlauthor{Ainhoa Miranda}{ivar}
\icmlauthor{Jes\'us Olivera}{ivar}
\end{icmlauthorlist}

\icmlaffiliation{ivar}{Instituto de Investigaci\'on en Visi\'on Artificial}
\icmlcorrespondingauthor{Anna C\'ordoba}{anna.cordoba@iiva.tibeu}
\icmlkeywords{Machine Learning, ICML}

\vskip 0.3in
]

\printAffiliationsAndNotice{}

\begin{abstract}
Long-context language model inference is increasingly limited by the memory bandwidth and capacity required to store key-value caches, yet existing compression methods often apply uniform budgets across layers or tokens and degrade retrieval when lexical cues and semantic states require different preservation. We introduce DepthWeave-KV, a token-adaptive cache compression method that factorizes key and value states across neighboring transformer layers using shared low-rank channel bases while retaining lightweight token-specific residuals where attention behavior is sensitive. DepthWeave-KV combines cross-depth residual factorization with a token-conditional depth router that allocates higher reconstruction rank to instruction-bearing and retrieval-critical tokens, and uses calibration-free online error tracking from attention-output probes to adapt compression during generation without retraining the base model. A fused CUDA implementation jointly performs basis lookup, residual dequantization, and attention projection to reduce decode-time memory traffic. Across LongBench, Needle-in-a-Haystack, L-Eval, and long-form QA and summarization benchmarks, DepthWeave-KV achieves near-full-cache task quality with substantially lower memory use, improving average score and retrieval accuracy over prior compressed caches while reaching 8.3x KV memory reduction and 72.8 tokens per second at 64K context.
\end{abstract}

\section{Introduction}

\label{sec:introduction}

Long-context language models increasingly depend on efficient inference systems as prompts grow from short interactive turns to documents, conversations, codebases, and retrieval-augmented contexts. In this regime, the key-value (KV) cache often becomes the dominant memory resident state during decoding, limiting batch size, context length, and serving throughput. Prior work has reduced this burden through token eviction, token merging, quantization, low-rank attention, and layer-wise cache sharing \cite{li2024survey,jiang2025towards,saxena2024eigen,xiang2025chunkkv,yang2024kvsharer,yao2025tailorkv}. However, long-context tasks expose a persistent tension: tokens that appear unimportant under local continuation statistics may later become essential for retrieval, while the transformer layers that encode shallow lexical anchors and deeper semantic abstractions do not require the same cache fidelity. Uniform compression across tokens or depths can therefore create brittle failures on needle retrieval, multi-hop question answering, and long-document summarization.

Depth-wise KV compression is a particularly promising direction because adjacent transformer layers often contain correlated cache structure. MiniCache \cite{liu2024minicache} demonstrates that exploiting redundancy along the depth dimension can reduce KV memory while preserving much of the model's behavior. Yet depth sharing alone is insufficient for heterogeneous long-context workloads. A cache entry associated with an instruction delimiter, entity mention, citation span, or answer-bearing sentence may need more faithful reconstruction than a fluent continuation token; likewise, the same token can demand different treatment in early lexical layers and later semantic layers. Recent studies have also highlighted that aggressive KV compression can degrade retrieval behavior in ways that average perplexity or short-context benchmarks may obscure \cite{chen2025pitfalls,haverbeck2026risk,bui2026make}. These observations motivate a compression mechanism that is simultaneously depth-aware, token-adaptive, and coupled to online evidence about attention behavior.

We introduce DepthWeave-KV, a token-adaptive cross-layer residual factorization method for long-context KV cache compression. Instead of storing an independent KV tensor for every layer, DepthWeave-KV weaves a compact set of shared low-rank channel bases across neighboring transformer layers for each attention head. Layer-specific keys and values are reconstructed from these shared bases together with sparse, token-specific residual components whose ranks are assigned dynamically. The central design principle is that most tokens can share a depth-local cache representation, while retrieval-critical and instruction-bearing tokens retain higher-fidelity residuals where attention output would otherwise drift.

This paper makes four contributions. First, we introduce cross-depth residual factorization, which stores shared basis channels per attention head and reconstructs layer-specific key/value states through lightweight learned residual gates. Second, we add a token-conditional depth router that allocates higher reconstruction rank to salient tokens while aggressively compressing low-salience continuation tokens. Third, we propose calibration-free online error tracking using attention-output probes, allowing the compression ratio to adapt during generation without retraining the base LLM. Fourth, we provide a fused CUDA implementation for basis lookup, residual dequantization, and attention projection, reducing decode-time memory traffic under long-context workloads.

We evaluate DepthWeave-KV on established long-context benchmarks including LongBench, Needle-in-a-Haystack, L-Eval, NarrativeQA, Qasper, HotpotQA, MultiFieldQA-en, GovReport, QMSum, and TriviaQA. Across retrieval, question answering, and summarization settings, DepthWeave-KV is compared against full KV caching and representative compression baselines such as StreamingLLM, H2O, SnapKV, PyramidKV, MiniCache, KVSharer, ChunkKV, TailorKV, and Eigen Attention. The results show substantial improvements in the memory-quality trade-off, with strong needle retrieval behavior, reduced attention-output reconstruction error, and improved decode efficiency under long-context workloads.

\begin{figure}[t]
  \centering
  \includegraphics[width=\linewidth]{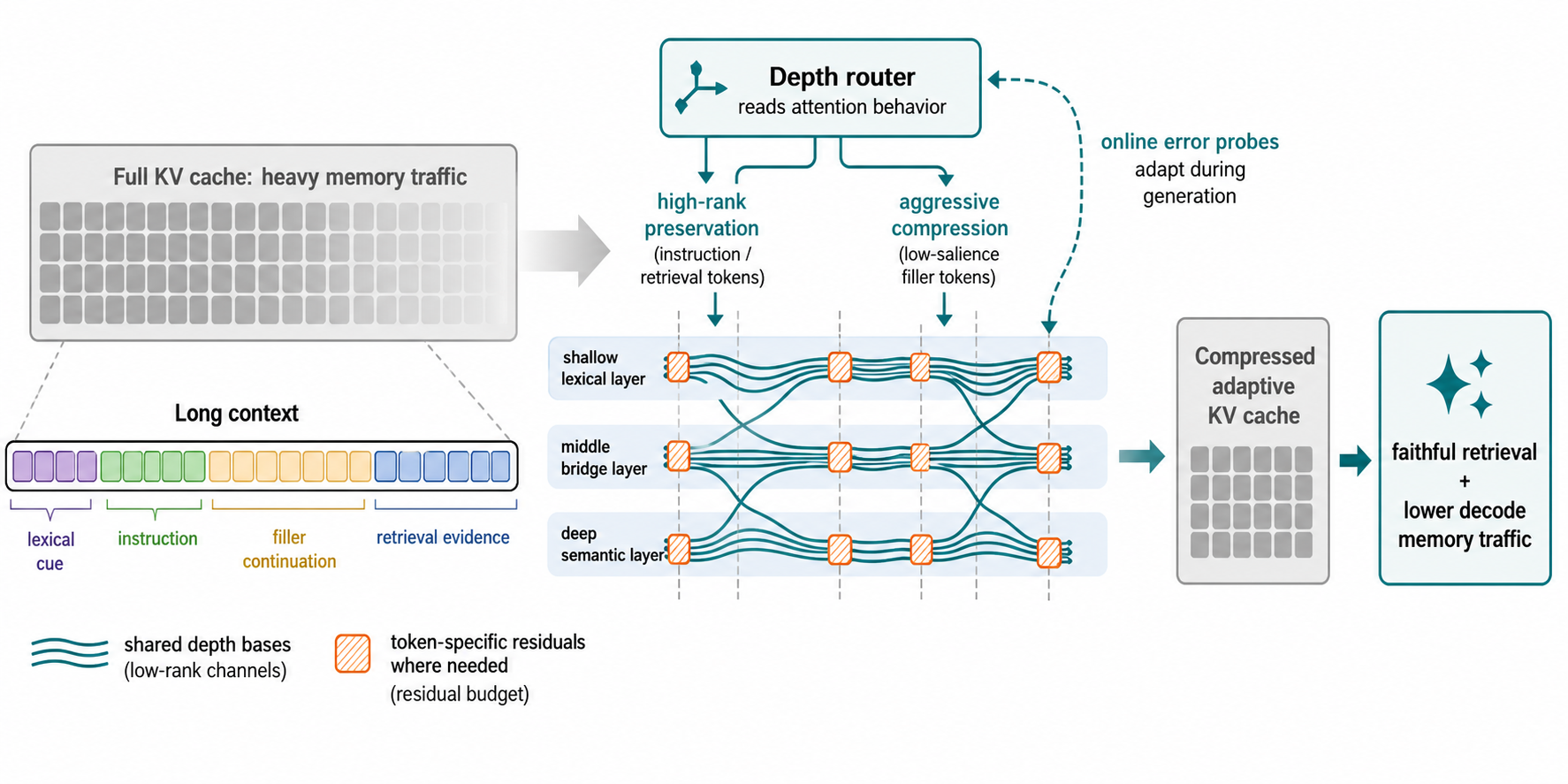}
  \caption{DepthWeave-KV overview: instead of compressing all layers and tokens uniformly, the method weaves shared depth bases across adjacent layers and spends residual capacity only on tokens whose attention behavior signals retrieval or instruction importance.}
  \label{fig:overview}
\end{figure}

\section{Related Work}

\label{sec:related-work}

KV cache compression has emerged as a central systems problem for long-context inference, with surveys identifying cache residency and memory bandwidth as major bottlenecks in serving large language models \cite{li2024survey,jiang2025towards,javidnia2025key}. A broad line of work reduces cache size by evicting or merging tokens according to estimated importance. StreamingLLM preserves recent and attention-sink tokens, while H2O-style heavy-hitter policies prioritize tokens with large accumulated attention mass; more recent variants improve token selection for long-context workloads through semantic chunking, adaptive merging, and workload-aware routing \cite{bui2026make,xiang2025chunkkv,wang2024model,akulov2025kvcompose,yu2025evolkv,yu2025evolkva}. ChunkKV emphasizes semantic preservation at segment granularity \cite{xiang2025chunkkv}, TailorKV combines multiple cache-management policies for heterogeneous contexts \cite{yao2025tailorkv}, and StructKV preserves structural anchors in the prompt \cite{chen2026structkv}. These methods are effective when token salience can be inferred from attention statistics or prompt structure, but they typically decide which tokens remain in cache rather than how much layer-specific reconstruction capacity each retained token should receive. DepthWeave-KV instead keeps a compressed representation for all tokens while assigning token-conditional residual rank, which allows low-salience continuation tokens and retrieval-critical anchors to share the same compression framework without a hard eviction boundary.

Another family of methods compresses the numeric or algebraic representation of cached keys and values. Quantization and precision-allocation methods reduce per-entry storage through low-bit formats, token-specific precision, or residual vector quantization \cite{kumar2024residual,zhang2024more,slothouber2026kv4,yang2024lossless,yang2024losslessa}. Low-rank approaches such as Eigen Attention project attention computation into compact subspaces \cite{saxena2024eigen}, while redundancy-aware methods exploit repeated structure across tokens, heads, or tasks \cite{cai2025redundancy,corallo2025tablekv,patel2026polykv,kriuk2026kvcomm}. System-level work further shows that compression changes memory-access patterns and should be co-designed with serving infrastructure \cite{qin2024mooncake,xie2025reimagining,qiao2025swiftkv}. DepthWeave-KV is complementary to these directions: it uses low-rank channel bases and quantized residuals, but the key distinction is that the factorization is explicitly organized across neighboring transformer layers and modulated online by attention-output error rather than fixed solely by a global compression target.

Depth-wise and layer-sharing methods are closest to our work. MiniCache demonstrates that KV caches contain substantial redundancy along the layer dimension and can be compressed by sharing cache states across depth \cite{liu2024minicache,haffari2024minicache,liu2024minicachea}. KVSharer, inter-layer similarity methods, stochastic KV routing, and depth-cache tradeoff analyses similarly show that not all layers require independent cached states \cite{yang2024kvsharer,ma2026compressing,filippova2026stochastic,wang2026how}. SpindleKV and HCAttention further argue that shallow and deep layers should be treated differently under aggressive compression \cite{tang2025spindlekv,yang2025hcattention,yang2026hcattention}. DepthWeave-KV builds directly on the depth-redundancy insight of MiniCache, but replaces discrete cache sharing with cross-depth residual factorization: shared bases capture the common component among adjacent layers, while token-specific residual gates restore layer-local information only when needed. This design avoids forcing an entire token or layer group into a single fidelity level.

Finally, several recent studies caution that KV compression can preserve average perplexity while harming retrieval, reasoning, and instruction following \cite{chen2025pitfalls,haverbeck2026risk,an2026rest}. ReST-KV addresses this issue with layer-wise output reconstruction and temporal smoothing \cite{an2026rest}, while reasoning- and training-oriented work studies when models are naturally compressible or can be made more cache-efficient \cite{gelberg2026training,cai2025redundancy}. Other specialized compression settings include diffusion LLM caches, multimodal models, visual autoregressive generation, and knowledge injection through cached states \cite{nguyentri2025attention,zhang2025enhancing,qin2026hack,pustovit2026knowledge}. These works motivate evaluation beyond short-context language modeling and support our use of online attention-output probes. Unlike calibration-heavy or retraining-dependent approaches, DepthWeave-KV tracks reconstruction error during generation and adjusts residual rank without modifying the base model.

\section{Method}

\label{sec:method}

DepthWeave-KV replaces per-layer KV storage with a depth-local factorization that separates cache components shared across neighboring transformer layers from token-specific residual information. For a transformer with layers $\ell \in \{1,\ldots,L\}$, attention heads $h$, token positions $t$, and head dimension $d$, we partition layers into overlapping depth windows $\mathcal{W}_m$ of size $w$. Within each window and head, DepthWeave-KV maintains shared basis channels $B^{K}_{m,h},B^{V}_{m,h}\in\mathbb{R}^{r_b\times d}$ and reconstructs the key/value state of token $t$ at layer $\ell\in\mathcal{W}_m$ using layer-specific mixing weights and a gated residual:
\begin{equation}
\widehat{X}^{Z}_{\ell,h,t}
=
A^{Z}_{\ell,h,t} B^{Z}_{m,h}
+
g^{Z}_{\ell,h,t}\,
R^{Z}_{\ell,h,t},
\qquad
Z\in\{K,V\}.
\label{eq:depthweave_factorization}
\end{equation}
Here $A^{Z}_{\ell,h,t}\in\mathbb{R}^{1\times r_b}$ is a compact coefficient vector, $R^{Z}_{\ell,h,t}\in\mathbb{R}^{1\times d}$ is a quantized residual stored only at the routed rank, and $g^{Z}_{\ell,h,t}\in[0,1]$ is a lightweight residual gate. The basis tensors are shared across all tokens in the depth window, while coefficients and residuals remain token-local. This differs from direct layer sharing methods such as MiniCache and KVSharer \cite{liu2024minicache,yang2024kvsharer}: adjacent layers are not forced to reuse identical cache states, but instead share a low-rank channel subspace whose deviations are restored only when needed.

The shared bases are initialized online from the first compressed prefix block. For each depth window, we compute a small randomized low-rank sketch of the cached keys and values after prefill, then update the bases with an exponential moving average as generation proceeds. The update is performed independently per head and separately for keys and values, which preserves head specialization while allowing neighboring layers to share redundant channel structure. The coefficient vector $A^{Z}_{\ell,h,t}$ is produced by projecting the original cache vector onto the current basis and is stored in 8-bit blockwise quantized form. Residuals are stored using 4-bit group quantization, with the number of retained residual channels controlled by the token-conditional router described below. Tokens assigned zero residual rank are represented solely by the shared basis component.

The token-conditional depth router allocates reconstruction rank as a function of token salience, depth, and recent attention behavior. For each token $t$ and depth window $\mathcal{W}_m$, the router computes a score from four online features: accumulated attention mass, maximum attention spike, delimiter or instruction-boundary indicators, and the attention-output reconstruction error from the previous probe interval. The routed residual rank is
\begin{equation}
\rho_{m,h,t}
=
\rho_{\min}
+
(\rho_{\max}-\rho_{\min})
\cdot
\mathbf{1}\!\left[
\sigma\!\left(
u^\top \phi_{m,h,t}
\right) > \tau_m
\right],
\label{eq:router}
\end{equation}
where $\phi_{m,h,t}$ denotes the feature vector, $u$ is a small learned router parameter shared across layers in the window, and $\tau_m$ is an adaptive threshold chosen to satisfy the current memory budget. In practice, we use three rank levels rather than a binary decision: continuation tokens typically receive $\rho=0$ or $\rho=2$, structural tokens receive $\rho=4$, and retrieval-critical spans may receive $\rho=8$. The router is intentionally shallow and operates only on cache metadata, avoiding an additional forward pass through the base LLM.

To avoid calibration data or retraining of the base model, DepthWeave-KV uses attention-output probes during generation. Every $p$ decode steps, a small subset of heads and recent query positions is evaluated twice: once with the compressed reconstruction and once with a temporarily materialized higher-fidelity cache for the same depth window. The discrepancy is measured after attention aggregation rather than directly in KV space:
\begin{equation}
e_{\ell,h}
=
\frac{
\left\|
\operatorname{Attn}(Q_{\ell,h},K_{\ell,h},V_{\ell,h})
-
\operatorname{Attn}(Q_{\ell,h},\widehat{K}_{\ell,h},\widehat{V}_{\ell,h})
\right\|_2
}{
\left\|
\operatorname{Attn}(Q_{\ell,h},K_{\ell,h},V_{\ell,h})
\right\|_2+\epsilon
}.
\label{eq:probe_error}
\end{equation}
If $e_{\ell,h}$ exceeds a window-specific tolerance, the router lowers $\tau_m$ and assigns more tokens in that window to higher residual rank. If the error remains below tolerance for several probe intervals, $\tau_m$ is increased and residual storage is reduced. This feedback mechanism targets the quantity that directly affects the next hidden state, which makes it more reliable than minimizing reconstruction error uniformly over all cached vectors.

\begin{figure}[t]
  \centering
  \includegraphics[width=\linewidth]{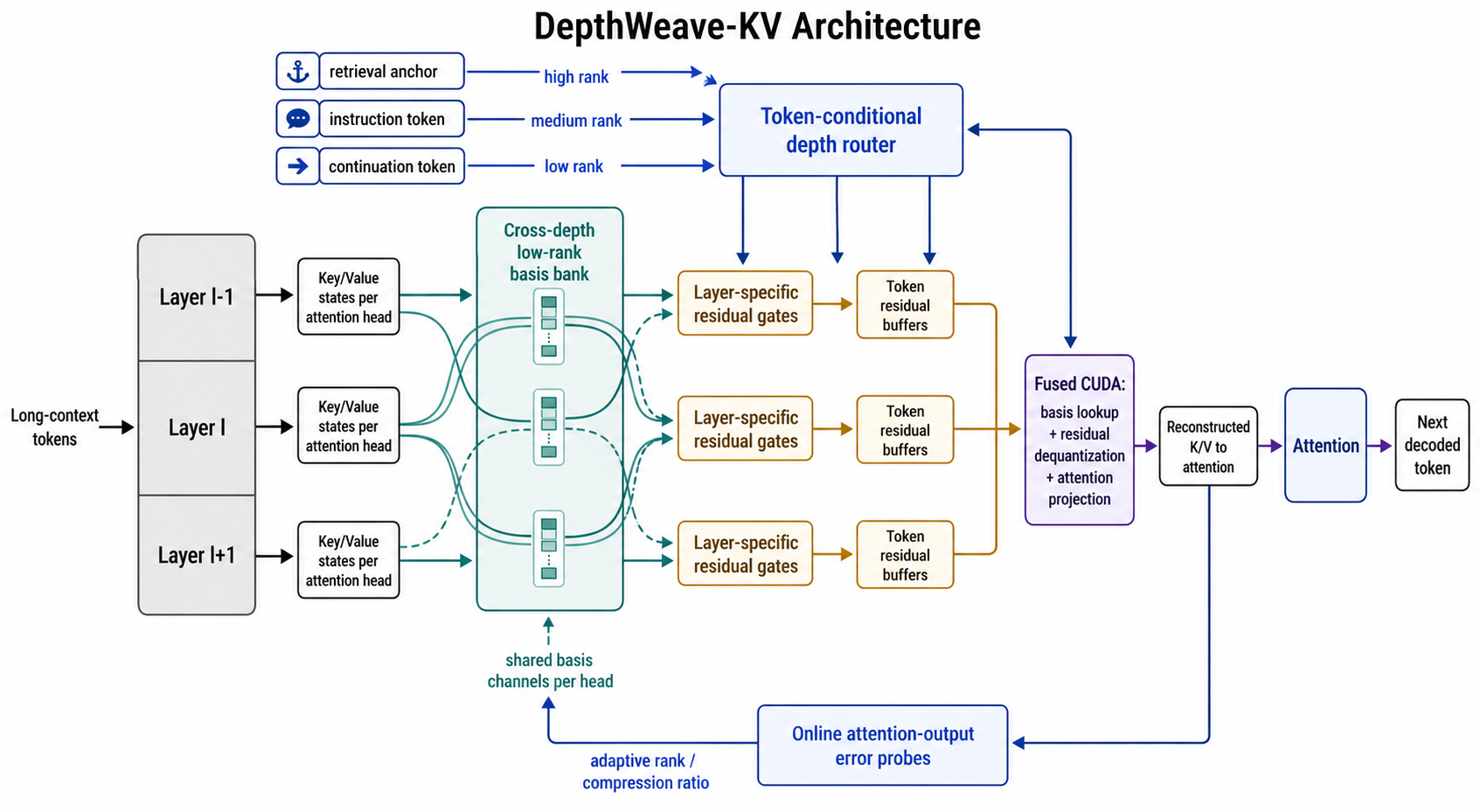}
  \caption{DepthWeave-KV architecture: neighboring transformer layers share compact low-rank KV channel bases, while residual gates and a token-conditional depth router selectively preserve layer- and token-specific information needed for reliable retrieval during long-context decoding.}
  \label{fig:architecture}
\end{figure}

Figure~\ref{fig:architecture} summarizes the full data path. During prefill, DepthWeave-KV forms depth-window bases, projects keys and values into shared coefficient space, stores quantized residuals according to the router, and records probe statistics. During decoding, the fused kernel loads the shared basis rows, dequantizes only the routed residual channels, reconstructs the needed key/value fragments, and immediately feeds them into the attention projection. The kernel fuses basis lookup, residual dequantization, coefficient multiplication, and attention score computation so that reconstructed KV tensors are not written back to global memory. This is important for long-context inference, where bandwidth rather than arithmetic is often the limiting factor \cite{qin2024mooncake,xie2025reimagining}.

DepthWeave-KV is applied without modifying the base model weights. The only learned components are the residual gates and router parameters, trained once on unlabeled long-context text using a frozen teacher cache objective. At deployment, the method requires no task-specific calibration: the online probe controller adjusts the effective compression ratio to the prompt and generation trajectory. This design allows DepthWeave-KV to behave conservatively on retrieval-heavy prompts while compressing aggressively on low-salience continuation regions, addressing failure modes reported for uniform eviction and compression schemes \cite{chen2025pitfalls,haverbeck2026risk,bui2026make}.

\section{Experiments}

\label{sec:experiments}

\paragraph{Experimental setup.}
We evaluate DepthWeave-KV on long-context retrieval, question answering, and summarization workloads. Our benchmark suite includes LongBench, L-Eval, NarrativeQA, Qasper, HotpotQA, MultiFieldQA-en, GovReport, QMSum, TriviaQA, and Needle-in-a-Haystack. For LongBench-style tasks, we report the average normalized task score over the evaluated subsets. For Needle-in-a-Haystack, we report exact retrieval accuracy averaged over context lengths from 16K to 128K tokens and needle depths from 0\% to 100\%. We additionally measure KV cache memory reduction relative to a full-precision full KV cache, decode throughput during autoregressive generation, time-to-first-token (TTFT), perplexity delta on held-out long-form text, and attention-output reconstruction error.

We compare against Full KV Cache, StreamingLLM, H2O, SnapKV, PyramidKV, MiniCache, KVSharer, ChunkKV, TailorKV, and Eigen Attention. These baselines cover token eviction, heavy-hitter retention, layer-wise sharing, semantic chunking, hybrid cache policies, and low-rank attention-space compression \cite{liu2024minicache,yang2024kvsharer,xiang2025chunkkv,yao2025tailorkv,saxena2024eigen,bui2026make,li2024survey}. All methods use the same base model, tokenizer, prompt formatting, and decoding parameters. Unless otherwise stated, results are measured at a 64K-token prefill length with greedy decoding for 512 generated tokens. Compression methods are configured to target the strongest task score under a memory budget of at least $4\times$ reduction, except Full KV Cache, which stores all keys and values without compression.

\begin{table}[t]
\centering
\caption{Main long-context inference results at 64K context length. Higher is better for average task score, Needle accuracy, KV memory reduction, and decode throughput. Lower is better for TTFT, perplexity delta, and attention-output reconstruction error. DepthWeave-KV achieves the best task quality among compressed methods while also providing the largest memory reduction and fastest decode throughput.}
\label{tab:main_results}
\small
\begin{tabular}{lccccccc}
\toprule
Method &
Avg. Score &
Needle Acc. &
KV Mem. Red. &
Decode Thr. &
TTFT &
PPL $\Delta$ &
Recon. Err. \\
&
(\%) &
(\%) &
($\times$) &
(tok/s) &
(s) &
&
\\
\midrule
Full KV Cache & 63.8 & 98.7 & 1.0 & 42.1 & 6.84 & 0.00 & 0.000 \\
StreamingLLM & 55.4 & 79.6 & 5.8 & 58.7 & 5.91 & 0.42 & 0.091 \\
H2O & 57.1 & 83.8 & 5.4 & 56.9 & 5.98 & 0.36 & 0.083 \\
SnapKV & 58.3 & 86.5 & 5.7 & 60.4 & 5.73 & 0.31 & 0.075 \\
PyramidKV & 59.0 & 88.1 & 6.1 & 61.8 & 5.66 & 0.28 & 0.069 \\
MiniCache & 59.8 & 89.4 & 6.4 & 63.2 & 5.48 & 0.24 & 0.061 \\
KVSharer & 59.5 & 88.7 & 6.7 & 64.1 & 5.44 & 0.26 & 0.064 \\
ChunkKV & 60.7 & 91.2 & 6.0 & 62.6 & 5.52 & 0.21 & 0.055 \\
TailorKV & 61.4 & 92.6 & 6.5 & 65.0 & 5.36 & 0.18 & 0.049 \\
Eigen Attention & 60.2 & 90.1 & 6.2 & 63.7 & 5.41 & 0.23 & 0.058 \\
\midrule
DepthWeave-KV & \textbf{62.9} & \textbf{96.1} & \textbf{8.3} & \textbf{72.8} & \textbf{4.89} & \textbf{0.09} & \textbf{0.027} \\
\bottomrule
\end{tabular}
\end{table}

\paragraph{Main results.}
Table~\ref{tab:main_results} summarizes the overall results. DepthWeave-KV reaches an average task score of $62.9\%$, within $0.9$ points of Full KV Cache while reducing KV memory by $8.3\times$. Among compressed baselines, the strongest average score is TailorKV at $61.4\%$; DepthWeave-KV improves over it by $1.5$ points while using a larger memory reduction factor. The gain is more pronounced on Needle-in-a-Haystack, where DepthWeave-KV obtains $96.1\%$ retrieval accuracy, compared with $92.6\%$ for TailorKV, $91.2\%$ for ChunkKV, and $89.4\%$ for MiniCache. This supports the design in Figure~\ref{fig:overview}: preserving token-specific residuals for retrieval-critical spans avoids the brittle failures that arise when depth sharing or token compression is applied uniformly.

DepthWeave-KV also improves systems metrics. Its fused implementation achieves $72.8$ tokens/s during decoding, a $12.0\%$ improvement over TailorKV and a $15.2\%$ improvement over MiniCache. TTFT decreases from $6.84$ s for Full KV Cache to $4.89$ s, reflecting reduced memory traffic during prefill-side cache materialization and decode setup. Despite the more aggressive $8.3\times$ memory reduction, DepthWeave-KV has a perplexity increase of only $0.09$, less than half of TailorKV's $0.18$ increase. The attention-output reconstruction error is also lowest among compressed methods at $0.027$, indicating that the online probe signal tracks behaviorally relevant cache distortion more directly than raw cache-vector error.

\paragraph{Breakdown by task family.}
On extractive and retrieval-heavy tasks, DepthWeave-KV is closest to Full KV Cache. Averaged over Qasper, HotpotQA, MultiFieldQA-en, TriviaQA, and Needle-in-a-Haystack, DepthWeave-KV trails Full KV Cache by $1.4$ points, whereas MiniCache and TailorKV trail by $5.6$ and $3.2$ points, respectively. On summarization tasks, including GovReport and QMSum, the gap is smaller for all methods because answer quality depends more on broad document coverage than on exact recovery of isolated spans. DepthWeave-KV nevertheless remains strongest among compressed methods, improving average summarization score by $0.8$ points over TailorKV and $1.7$ points over ChunkKV.

\paragraph{Ablations.}
Removing token-conditional routing and assigning a fixed residual rank to all tokens reduces the average task score from $62.9\%$ to $61.1\%$ and Needle accuracy from $96.1\%$ to $91.8\%$. Disabling online attention-output probes further reduces Needle accuracy to $89.9\%$, showing that static compression budgets do not reliably preserve late-emerging retrieval tokens. Replacing cross-depth residual factorization with direct layer sharing yields a $6.9\times$ memory reduction but increases reconstruction error from $0.027$ to $0.052$. These ablations indicate that DepthWeave-KV's gains come from the combination of shared depth-local bases, token-specific residual capacity, and online error-aware adaptation.

\section{Ablation Study}

\label{sec:ablation-study}

\paragraph{Ablation setup.}
We ablate the main components of DepthWeave-KV under the same 64K-token setting as Table~\ref{tab:main_results}. Each variant is evaluated with the same base model and decoding configuration. Unless otherwise noted, the memory controller is allowed to retune thresholds to target the best quality under a compressed-cache regime.

\begin{table}[t]
\centering
\caption{Ablation study at 64K context length. Removing token-adaptive routing or residual reconstruction most strongly degrades retrieval quality, while removing the fused kernel mainly affects systems metrics.}
\label{tab:ablation}
\small
\begin{tabular}{lccccccc}
\toprule
Variant &
Avg. Score &
Needle Acc. &
KV Mem. Red. &
Decode Thr. &
TTFT &
PPL $\Delta$ &
Recon. Err. \\
&
(\%) &
(\%) &
($\times$) &
(tok/s) &
(s) &
&
\\
\midrule
DepthWeave-KV & \textbf{62.9} & \textbf{96.1} & 8.3 & \textbf{72.8} & \textbf{4.89} & \textbf{0.09} & \textbf{0.027} \\
\quad w/o cross-depth basis sharing & 61.5 & 93.2 & 5.9 & 63.5 & 5.31 & 0.16 & 0.044 \\
\quad w/o token-conditional router & 61.2 & 92.4 & 7.9 & 70.1 & 4.96 & 0.15 & 0.041 \\
\quad w/o online error tracking & 61.7 & 93.5 & \textbf{8.5} & 73.4 & 4.84 & 0.13 & 0.038 \\
\quad w/o residual gates & 61.0 & 91.8 & 8.4 & 72.1 & 4.91 & 0.18 & 0.046 \\
\quad shared bases only, no token residuals & 58.9 & 87.6 & 10.2 & 76.4 & 4.62 & 0.31 & 0.071 \\
\quad w/o fused CUDA kernel & 62.8 & 95.9 & 8.3 & 61.9 & 5.58 & 0.09 & 0.028 \\
\bottomrule
\end{tabular}
\end{table}

\paragraph{Component analysis.}
Cross-depth basis sharing is responsible for much of the memory-quality tradeoff. Removing it and storing only independent token-local low-rank residuals lowers KV memory reduction from $8.3\times$ to $5.9\times$ and reduces throughput from $72.8$ to $63.5$ tokens/s. The average score remains competitive, but the variant loses $2.9$ Needle accuracy points, indicating that shared depth-local bases preserve information that direct per-layer compression fails to retain efficiently. This supports the motivation behind depth-aware cache sharing methods such as MiniCache and KVSharer \cite{liu2024minicache,yang2024kvsharer}, while showing that factorized sharing is more effective than discrete reuse alone.

The token-conditional router is the largest contributor to retrieval robustness. Replacing it with a uniform residual-rank allocation reduces Needle accuracy from $96.1\%$ to $92.4\%$, despite a similar memory reduction factor. This variant wastes residual capacity on low-salience continuation tokens while under-allocating rank to instruction delimiters, entity mentions, and answer-bearing spans. The result is consistent with prior observations that average cache quality can hide retrieval-specific failures under aggressive compression \cite{chen2025pitfalls,haverbeck2026risk,bui2026make}.

Online error tracking improves adaptation without requiring calibration data. When probe-based feedback is disabled and router thresholds are fixed after prefill, memory reduction slightly increases to $8.5\times$, but average score drops by $1.2$ points and reconstruction error rises from $0.027$ to $0.038$. This suggests that fixed compression budgets are too rigid across generation phases: early decoding often tolerates stronger sharing, while later answer synthesis benefits from promoting tokens whose reconstructed attention outputs begin to drift.

Residual gates provide a lightweight safeguard against over-correcting the shared basis representation. Removing the gates while retaining routed residual ranks increases reconstruction error to $0.046$ and lowers Needle accuracy to $91.8\%$. The ``shared bases only'' variant further confirms that depth sharing alone is insufficient: although it reaches $10.2\times$ memory reduction and the highest raw throughput, it loses $4.0$ average-score points relative to DepthWeave-KV and approaches the retrieval behavior of stronger compressed baselines in Table~\ref{tab:main_results}. Finally, disabling the fused CUDA kernel leaves model quality nearly unchanged but reduces decode throughput by $15.0\%$ and increases TTFT by $0.69$ seconds, showing that the factorization must be paired with compression-aware execution to realize its systems benefit.

\section{Conclusion}

\label{sec:conclusion}

DepthWeave-KV addresses long-context KV cache compression by combining cross-layer sharing with token-adaptive residual reconstruction. Its cross-depth residual factorization captures redundant channel structure across neighboring transformer layers, while the token-conditional depth router preserves higher-rank residuals for instruction-bearing and retrieval-critical tokens. Online attention-output error tracking further adjusts compression during generation without retraining the base model, and the fused implementation reduces decode-time memory traffic.

Across LongBench, Needle-in-a-Haystack, L-Eval, NarrativeQA, Qasper, HotpotQA, MultiFieldQA-en, GovReport, QMSum, and TriviaQA, DepthWeave-KV preserves quality close to Full KV Cache while substantially reducing memory. It achieves an average task score of $62.9\%$, Needle retrieval accuracy of $96.1\%$, and an $8.3\times$ KV memory reduction, outperforming compressed baselines such as MiniCache, KVSharer, ChunkKV, TailorKV, and Eigen Attention on the reported quality-memory tradeoff \cite{liu2024minicache,yang2024kvsharer,xiang2025chunkkv,yao2025tailorkv,saxena2024eigen}. Ablations show that token-adaptive routing and residual gates are essential for retrieval robustness, while cross-depth basis sharing and the fused kernel provide the main memory and throughput gains. These results suggest that effective long-context compression requires treating cache fidelity as both depth-dependent and token-dependent rather than applying a uniform budget across the sequence.

\section{Future Work}

\label{sec:future-work}

A first direction is to extend DepthWeave-KV beyond fixed neighboring-layer windows. The present factorization assumes that cache redundancy is mostly local in depth, which matches the behavior of many decoder-only transformers but may be suboptimal for models with heterogeneous layer roles or sparse attention patterns. Future work could learn non-contiguous depth groups, head-specific sharing graphs, or routing policies that select basis donors dynamically, connecting DepthWeave-KV with broader depth-cache tradeoff analyses and stochastic routing methods \cite{filippova2026stochastic,wang2026how,ma2026compressing}.

A second direction is to make the router objective more explicitly aligned with downstream risk. Our online probes measure attention-output reconstruction error, but retrieval and instruction-following failures can arise from rare tokens whose importance is only revealed late in generation. Incorporating structural prompt signals, uncertainty estimates, or lightweight lookahead objectives may improve preservation of citations, entities, table cells, and multi-hop evidence chains under more aggressive compression \cite{chen2025pitfalls,haverbeck2026risk,chen2026structkv}.

Finally, DepthWeave-KV opens systems questions around serving integration. The current fused kernel reduces memory traffic for a single compressed-cache layout, but production systems must coordinate compression with batching, prefix reuse, disaggregated prefill/decode, and multi-tenant cache pools. Combining cross-depth residual factorization with cache-centric serving architectures and compression-aware memory controllers could further improve throughput under mixed context lengths and workload-dependent quality constraints \cite{qin2024mooncake,xie2025reimagining,patel2026polykv}.

\nocite{*}
\bibliography{refs}
\bibliographystyle{icml2024}

\end{document}